\pdfoutput=1

\documentclass[11pt]{article}

\usepackage[preprint]{acl}

\usepackage{times}
\usepackage{latexsym}
\usepackage{booktabs}
\usepackage{tcolorbox}
\usepackage{lipsum} 
\usepackage{xcolor} 
\usepackage{multirow} 
\usepackage[T1]{fontenc}

\usepackage[utf8]{inputenc}

\usepackage{microtype}

\usepackage{inconsolata}

\usepackage{graphicx}

\title{Psychological Counseling Cannot Be Achieved Overnight: Automated Psychological Counseling Through Multi-Session Conversations}

\author{
  Bichen Wang\thanks{\ \ Equal contribution.},
  Junzhe Wang\footnotemark[1],
  Xing Fu,
  Yixin Sun,
  Bing Qin ,
  Yanyan Zhao\thanks{\ \ Corresponding author.} \\
   Research Center for Social Computing and Information Retrieval\\
 Harbin Institute of Technology, China\\
 {junzhewang,bichenwang}@ir.hit.edu.cn \\
}

\begin{document}
\maketitle
\begin{abstract}

In recent years, Large Language Models (LLMs) have made significant progress in automated psychological counseling. However, current research focuses on single-session counseling, which doesn't represent real-world scenarios. In practice, psychological counseling is a process, not a one-time event, requiring sustained, multi-session engagement to progressively address clients' issues. To overcome this limitation, we introduce a dataset for \textbf{Mu}lti-\textbf{S}ession \textbf{Psy}chological Counseling Conversation Dataset (\textbf{MusPsy}-Dataset). Our \textbf{MusPsy}-Dataset is constructed using real client profiles from publicly available psychological case reports. It captures the dynamic arc of counseling, encompassing multiple progressive counseling conversations from the same client across different sessions. Leveraging our dataset, we also developed our MusPsy-Model, which aims to track client progress and adapt its counseling direction over time. Experiments show that our model performs better than baseline models across multiple sessions.

\end{abstract}

\begin{figure}[t]
  \centering
  \includegraphics[width=1.05\columnwidth]{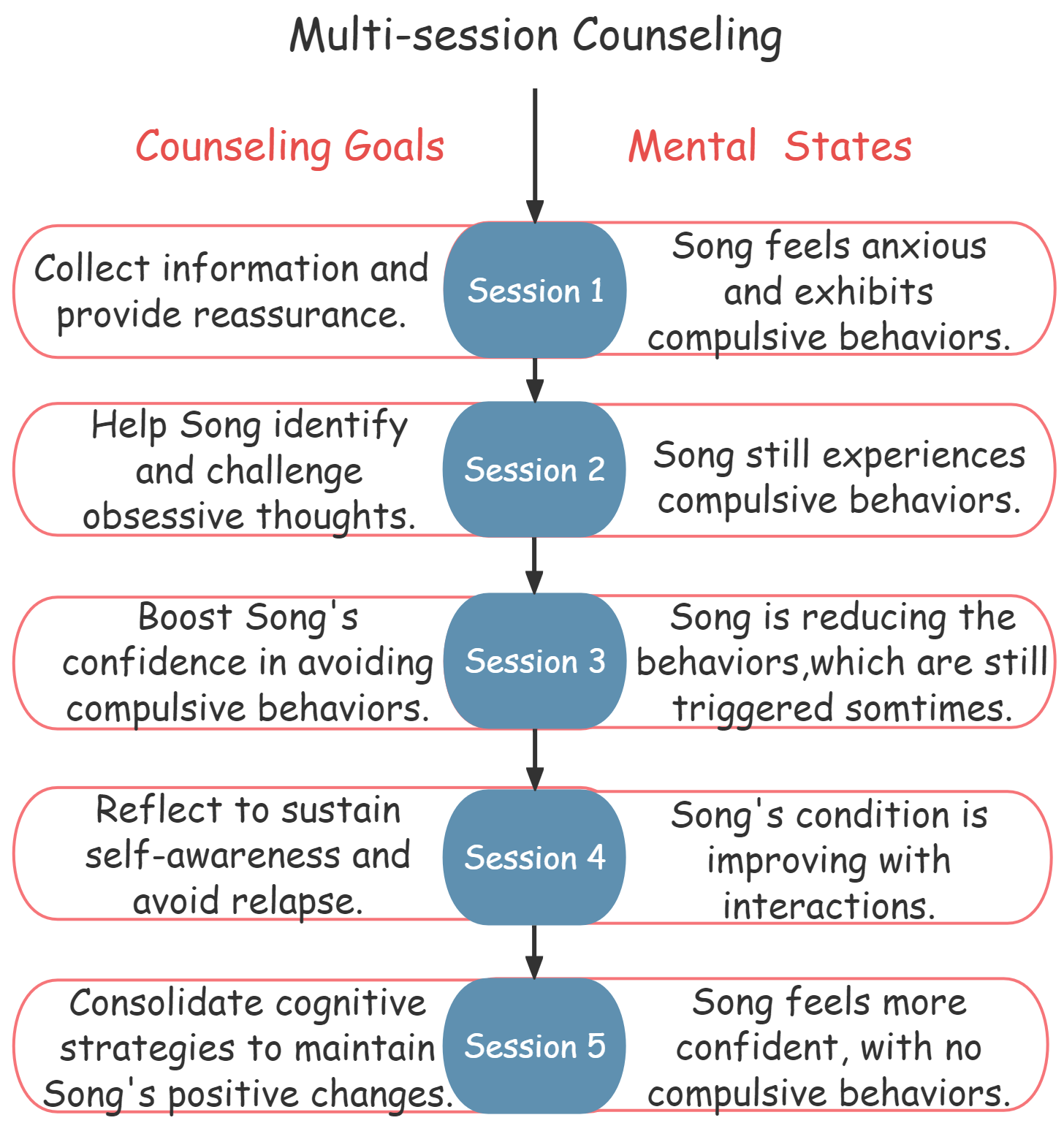}
  \caption{An real example of multi-session counseling involves a client demonstrating psychological improvement over five sessions. Each session targets specific goals, and the client's mental state gradually improves, an outcome that would be unachievable in a single-session setting.}
  \label{fig:intro}
\end{figure}

\section{Introduction}
In today’s society, individuals face increasing levels of psychological pressure~\cite{world2024global}, and the demand for mental health support is continuously surging due to the fast-paced nature of modern life~\cite{samji2022mental}. Consequently, the scarcity of accessible mental health services~\cite{grant2018three} has driven researchers to explore computational technologies for automated psychological counseling.

Previous research has advanced the modeling of psychological theory and the creation of various psychological counseling conversations, yielding positive results~\cite{lee2024cactus, qiu-etal-2024-smile, na2024cbt}. However, these researches are limited by their single-session format. Unlike casual chats, psychological counseling, such as Cognitive Behavioral Therapy (CBT), often needs to span multiple sessions~\cite{craske2010cognitive}. As shown in Figure \ref{fig:intro}, clients experience gradual psychological changes through counselor-guided multi-session counselings. During these sessions, counselors serve as navigators, dynamically recalibrating counseling goals based on the evolving dynamic mental states of their clients~\cite{baur2024content, dobson2021handbook}. This highlights the importance of developing automated models capable of sustaining coherent multi-session counseling conversations.

To address these limitations, our research introduces the integration of a multi-session paradigm into an automated psychological counseling framework. Unlike previous research that constructs isolated sessions, our work attempts to model multiple sessions to capture counseling progression. This approach accounts for the dynamically changing states of clients and allows for continuous adjustments to the counselors' goals. Consequently, we can cumulatively build client rapport, address clients' problems, and support long-term psychological improvement.

In general, we develop \textbf{MusPsy}-Dataset, a \textbf{Mu}lti-\textbf{S}ession \textbf{Psy}chological Counseling Conversation Dataset based on CBT using LLM. 
To ensure authenticity and quality, we collect client profiles and counseling goals spanning multiple sessions for the same client from publicly available psychological counseling case reports to generate multi-session counselings. Since LLMs struggle to generate coherent multi-session counselings in a single pass, we adopt a top-down approach: first, we generate multiple short seed conversations to guide the multiple conversation's flow; then, we elaborate this seed conversations into complete, consistent, and goal-driven long counseling conversations across sessions. When generating these complete conversations based on the seed conversations, we enhanced them by incorporating counseling fragments from our case reports and by using a carefully designed prompt. To simulate human counselor workflow and address long-context conversation challenges, we have designed a process to extract session-level memories from the counseling sessions.

We train an automated psychological counseling model using the MusPsy-Dataset to conduct multi-session counseling. We validate our model's effectiveness across multiple psychological metrics, including basic indicators, the Working Alliance Inventory (WAI), and the Positive and Negative Affect Schedule (PANAS).
\begin{itemize}
    \item We introduce a novel integration of multi-session conversation into automated counseling, enabling more longitudinal modeling of counseling progress.
    \item We introduce the MusPsy-Dataset, a novel multi-session psychological counseling conversation dataset based on CBT.
    \item We develop a multi-session counseling model, MusPsy-Model, trained on the MusPsy-Dataset, demonstrating its ability across multiple psychological metrics.
\end{itemize}

\section{Related Work}
\subsection{Cognitive Behavioral Therapy}
Cognitive Behavioral Therapy (CBT) has long been recognized as an effective intervention for individuals struggling with depression and anxiety~\cite{beck2020cognitive}. Previous researches describe how people with these disorders often develop negative, irrational thoughts that reinforce detrimental beliefs about themselves, others, and the world. To disrupt this cycle, CBT focuses on identifying and challenging these automatic thoughts and core beliefs ~\cite{longmore2007we}. CBT is a structured, multi-session process that cannot be accomplished in a single session~\cite{hayes2018process}. During CBT sessions, counselors first assist clients in recognizing unhelpful thoughts. They then guide clients in challenging and correcting these distortions using various CBT techniques, which ultimately help in reconstructing more positive automatic thoughts and beliefs over time~\cite{fenn2013key}. This counseling process is vital for promoting mental health and well-being.

\subsection{Automated Psychological Counseling}
Eliza, the pioneering system, used a rule-based approach for Rogerian counseling ~\cite{rogers1952client, weizenbaum1966eliza}. However, progress in data-driven research on Automated Psychological Counseling has been hindered by the limited availability of data. The creation of public, real, large-scale datasets for counseling remains impossible due to privacy concerns and the protection of vulnerable groups. However, the advent of LLMs has enabled the generation of synthetic data for counseling datasets. For instance, the SMILE dataset~\cite{qiu-etal-2024-smile}, an improvement over PsyQA~\cite{sun-etal-2021-psyqa}, includes multiple rounds of counseling dialogues and is used by CBT-LLM~\cite{qiu-etal-2024-smile}. CPsyCoun extracts counseling data from public psychology reports without requiring specific domain expertise ~\cite{zhang-etal-2024-cpsycoun}. Healme~\cite{xiao-etal-2024-healme} focuses on optimizing CBT guidance through LLMs, while CACTUS~\cite{lee2024cactus} emphasizes the construction of more complete single-turn CBT processes.

Compared to these efforts, our work incorporates multi-session dynamics into automated psychological counseling. We introduce a novel dataset and model.

\begin{figure*}[t]
  \includegraphics[width=1.0\textwidth]{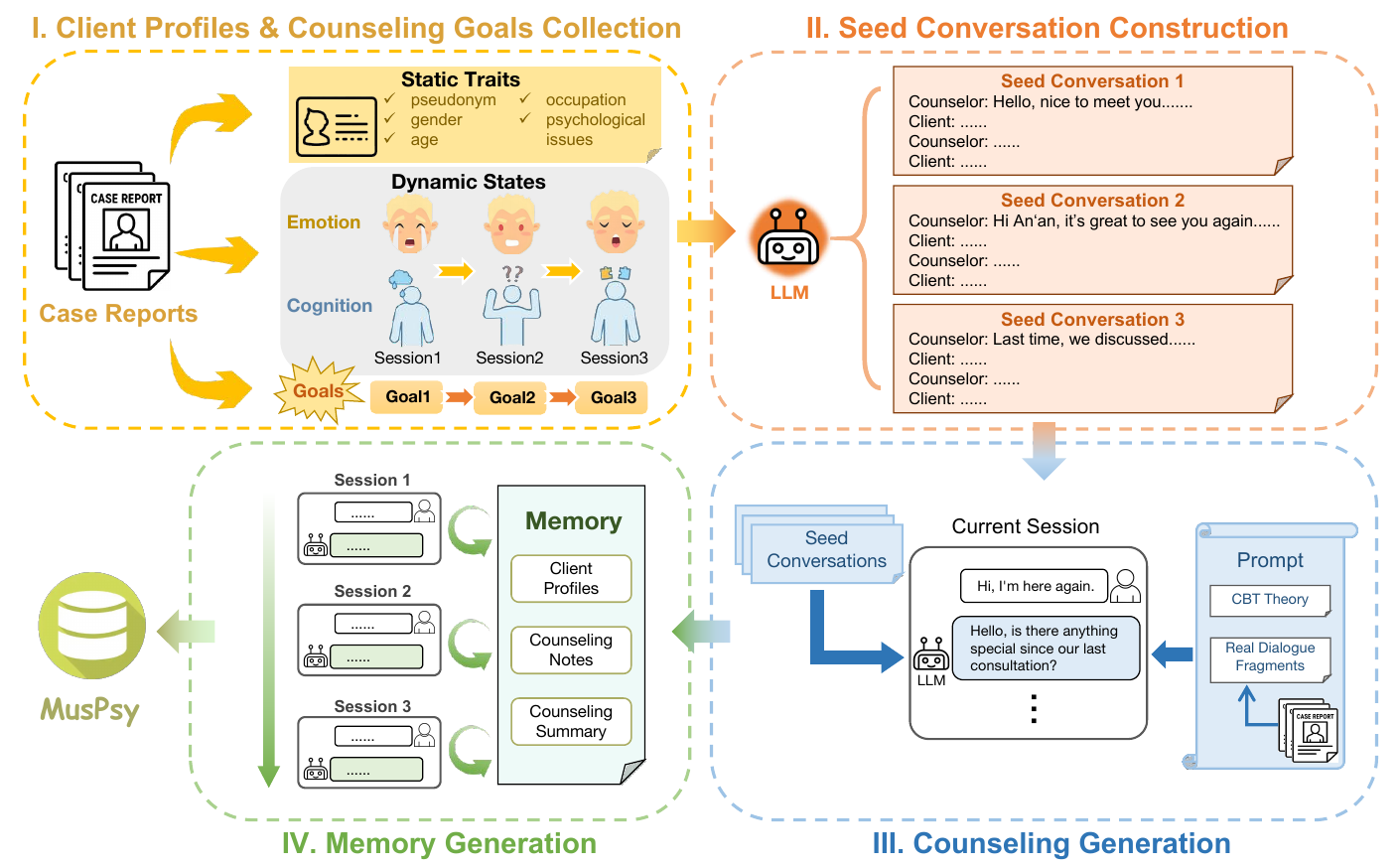}
  \centering
  \caption{This figure illustrates the process of constructing the MusPsy-Dataset. It includes four parts: Client Profiles \& Counseling Goals Collection, Seed Conversation Construction, Counseling Generation, and Memory Generation. Through these four steps, we obtained a high-quality MusPsy-Dataset.}
  \label{fig:data}
\end{figure*}

\section{MusPsy-Dataset: A Multi-Session Psychological Counseling Conversation Dataset}

We present the construction process of our MusPsy-Dataset. As shown in Figure \ref{fig:data}, in The construction of our dataset involves four key stages: \textbf{Client Profiles \& Counseling Goals Collection, Seed Conversation Construction,  Counseling Generation, and Memory Generation}.

\subsection{Client Profiles \& Counseling Goals Collection}

To support multi-session automated counseling, constructing realistic and context-aware client profiles is essential. While previous studies often rely on synthetic client features focusing only on static traits—such as age, gender, and presenting problems—they typically lack the dynamic evolution of a client's mental state across multiple sessions, resulting in limited support for long-term counseling simulations.

In our work, we address this issue by gathering multi-session counseling client profiles from credible psychological case reports published in academic journals or professional books, ensuring both authenticity and continuity. While these reports exclude full session transcripts for privacy reasons, they document the client’s state and achieve counseling goals per session. Unlike prior methods that only capture initial background, our profiles cover both static traits and dynamic states, as well as the counselor's planned counseling goals before each session.

\begin{itemize}
\item \textbf{Static Traits}: These include pseudonym, gender, age, occupation, and initial psychological issues reported during intake.
\item \textbf{Dynamic States}: These capture session-by-session changes, including the client’s evolving recent life events and emotional and cognitive state.
\item \textbf{Counseling Goals}: Counseling Goals define the specific direction and content of the counselor's intended session.
\end{itemize}

\begin{figure}[t]
  \includegraphics[width=0.5\textwidth]{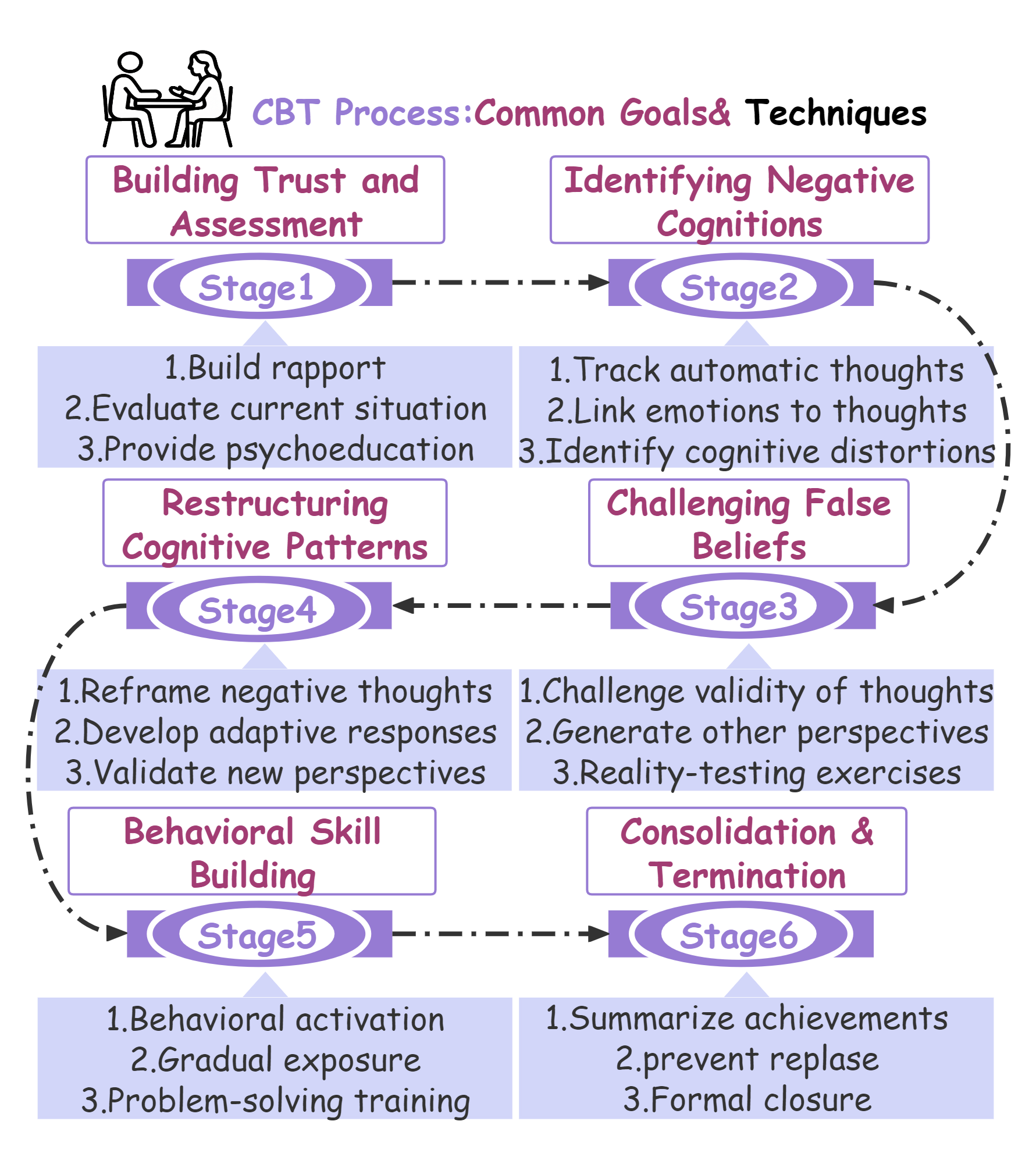}
  \caption{Multi-session CBT generally follows a paradigm in which the counselor typically adheres to this process, integrating it with the client's actual situation to set more specific and actionable counseling goals and use counseling techniques~\cite{beck2011cognitive}.}
  \label{fig:cbt_stage}
\end{figure}

Client profiles are composed of static traits and dynamic states. Counseling goals are not included in client profiles because they usually involve more than just the client. Figure \ref{fig:cbt_stage} illustrates typical multi-session CBT goals and methods, defining the counselor's intended session direction and content.

We use GPT-4o with human validation to extract structured information from case reports. See Appendix~\ref{sec:client_profile} for details on client profile and counseling goal collection.

\subsection{Seed Conversation Construction}
Directly generating all complete sessions at once suffers from limited depth due to LLM context and instruction-following constraints. Conversely, generating sessions separately often produces repetitive phrasing and weak continuity, as the LLM lacks awareness of counseling progression. This highlights the need for different generation methods that ensure both coherence and contextual richness across counseling sessions.

\paragraph{Generating Seed Conversations for Coherent Multi-Session Generation:} As shown in Figure \ref{fig:data}, to ensure coherence across multi-session interactions, we introduce an intermediate step: constructing seed conversations before generating full counseling conversations. Seed conversations refer to short and concise multi-session conversations generated by the LLM simultaneously, which, despite their brevity, possess high coherence due to the concurrent generation.

To ensure coherence, we input multiple client profiles and their corresponding goals together. We instruct the LLM to generate coherent seed conversations, each containing only 3 to 4 turns and about 100-200 tokens of concise communication aligned with the session's goal. For details on seed conversation construction, see Appendix~\ref{sec:seed_conver}.

\subsection{Counseling Generation}
Given initial multi-session conversations, we aim to generate complete, extended counselings that are contextually coherent across sessions and linguistically similar to real counselors.
\paragraph{Session-Level Expansion with Contextual Constraints:} During generation, to ensure contextual coherence, we prompt the LLM to generate each complete counselings based on the seed conversation while providing the preceding seed conversations. This encourages it to create contextually linked conversations. This design allows us to preserve temporal dependencies and emotional progression across multiple sessions, ensuring a realistic and coherent multi-session simulation.

\paragraph{Prompting with Few-Shot Guidance:} We design prompts to guide the generation process by combining descriptions of core CBT techniques with the LLM's own understanding of CBT to expand the seed conversations. We find some reports that provided anonymized, incomplete counseling fragments. We utilize real counseling fragments extracted from these reports as examples corresponding to the relevant client profile stages. These samples are embedded into the prompt to guide the LLM's tone, structure, and counseling techniques, resulting in more authentic and contextually appropriate outputs.

We use an efficient single-agent script-based approach that significantly lowers computational cost and inference time while keeping dialogue coherent and sound, as shown in prior single-session research~\cite{lee2024cactus}. Here, one LLM generates the entire multi-turn dialogue in a single pass, based on the user profile and the initial session conversation. See Appendix~\ref{sec:gen_conver} for counseling generation details.

\subsection{Memory Generation}
Additionally, long-term multi-session counseling can easily exceed the available context window and struggle to maintain consistency. To support consistent and context-aware multi-session counseling, we construct structured memory representations for each client.

Unlike general chat memory, the design of counseling memory needs to reflect clinical requirements. To meet this challenge, we introduce an external memory module that captures key session-level information, enabling us to maintain coherence and continuity over extended counseling sessions. Inspired by documentation standards in CBT practice \citep{bemister2011updated, lawlor2014digital}, we design a multi-part memory structure specifically tailored for multi-session counseling.

\begin{itemize}
\item \textbf{Client Profiles:} Includes client profiles and personal traits.

\item \textbf{Counseling Notes:} Captures key session-level information, including counselor observations, session goals, and counseling assignments during multiple counseling sessions.

\item \textbf{Counseling Summary:} A brief summary of the entire series of sessions.
\end{itemize}

At the end of each counseling session, we summarize the client's conversation into this memory structure. This approach allows the model to update and retain crucial information without needing to store the complete counseling transcripts. As a result, subsequent sessions can maintain continuity and simulate a counselor's cross-session tracking and planning abilities. For more details on memory generation, please refer to Appendix~\ref{sec:mem_gen}.

\section{Data Statistics and Evaluation}

\subsection{Data Statistics}
We provide an overview of the basic statistics of our dataset in Table \ref{tab:basic_metrics}. The MusPsy-Dataset consists of 1,400 clients' multi-turn counselings constructed based on CBT theory, with each session averaging 28.55 turns and 26.99 words per turn. On average, each client has 6.17 counseling sessions. This dataset scale is sufficient to support the construction of our new multi-session counseling model.
\begin{table}[ht]
\centering
\begin{tabular}{ll}
\hline
\textbf{Item} & \textbf{Value} \\
\hline
Client Profiles for Training & 1,400 \\
Client Profiles for Testing & 100 \\
Average Sessions Per Client & 6.17 \\
Average Turns Per Session & 28.55 \\
Average Words per Turn & 26.99 \\
Average Tokens Per Client & 5693.27 \\
Average Tokens Per Turn & 33.23 \\
\hline
\end{tabular}%
\captionof{table}{Data statistics of our MusPsy-Dataset}
\label{tab:basic_metrics}
\end{table}

\subsection{Competing Datasets}

We introduce several previous works for comparison, including the SMILE dataset~\cite{qiu-etal-2024-smile}, a multi-round mental health conversation dataset based on PsyQA; and the CACTUS dataset~\cite{lee2024cactus}, another multi-round CBT dataset for mental health. We also include the SimPsyDial~\cite{qiu2024interactive} and CPsyCoun~\cite{zhang-etal-2024-cpsycoun} datasets. We evaluate data quality and model performance as reported in these studies within our evaluation framework. These works are similar to ours in constructing mental health counseling datasets and providing mental support models.

\subsection{Dataset Quality Evaluations}

We randomly select 100 counseling examples per dataset. Two psychology experts perform manual evaluations, and GPT-4o does automated ones. We focus on four basic metrics and counselor-client working alliance, using it as a consistent measure across diverse counseling data \cite{horvath2001alliance}. Research indicates working alliance predicts counseling outcomes \cite{horvath1991relation, horvath2011alliance}. All metrics are rated 1-5.

We evaluate our dataset using basic metrics \cite{munder2010working} across four dimensions: \textbf{Helpfulness:} the practical utility of counselor explanations and information; \textbf{Empathy:} the counselor's ability to understand and share client feelings; \textbf{Guidance:} the availability and specificity of practical advice; and \textbf{Coherence:} the logical consistency in the conversation. We also use the 12-item Working Alliance Inventory (WAI) \cite{munder2010working}, which assesses three dimensions: \textbf{Goal Agreement:} mutual agreement on our counseling goals; \textbf{Task Agreement:} agreement on the methods used; and \textbf{Emotional Bond:} mutual trust, confidence, and liking. Each dimension in the WAI is assessed by three questions, and we report the average score for each dimension. For details on the evaluations, see Appendix \ref{sec:data_eval}.

\begin{table*}[ht!]
\centering
\begin{tabular}{c c ccccc|cccc}
\toprule
\textbf{Evaluator} & & \textbf{Hel.} & \textbf{Emp.} & \textbf{Gui.} & \textbf{Coh.} & \textbf{Avg.} & \textbf{Task.} & \textbf{Bond.} & \textbf{Goal.}& \textbf{Avg.}  \\
\midrule
\multirow{6}{*}{GPT-4o}&SMILE & 4.05 & 4.49 & 3.98 & 4.43&4.23  & 3.60 & 4.35&3.61&3.85\\
      &CACTUS & 4.62 & 4.86 & 4.56 & \textbf{4.99} &4.76      &3.56&4.56&3.66&3.92\\
      &SimPsyDial & 4.58 & 4.90 & 4.51 & 4.99 & 4.75 & 3.77 & 4.31 & 3.56 & 3.97 \\
      &CPsyCoun & 4.23 & 4.43 & 4.09 & 4.87 & 4.41 & 3.56 & 4.30 & 3.71 & 3.86 \\
      &\textbf{MusPsy} &\textbf{4.99} & \textbf{4.98} & \textbf{4.98} & \textbf{4.99}&\textbf{4.98} & \textbf{4.11}&\textbf{4.69}&\textbf{4.44}&\textbf{4.41}\\
    \midrule
    \multirow{6}{*}{Human}&SMILE& 3.84 & 3.90 & 3.57 & 3.77&3.77&3.13&4.22&3.14&3.49\\ 
      &CACTUS& 4.49 & 4.59 & 4.18 & 4.65& 4.47 &3.60&4.65&3.49&3.92\\ 
      &SimPsyDial & 3.73 & 4.56 & 4.32 & 4.77 & 4.34& 3.36 & 4.32 & 3.58 & 3.75 \\
      &CPsyCoun & 3.60 & 3.83 & 4.17 & 3.73 & 3.83 & 3.08 & 4.02 & 3.43 & 3.51 \\
      &\textbf{MusPsy} & \textbf{4.89} & \textbf{4.82} & \textbf{4.70} & \textbf{4.88}& \textbf{4.82}&\textbf{4.20}&\textbf{4.72}&\textbf{4.14}&\textbf{4.35}\\ 
\bottomrule
\end{tabular}%
\caption{Data Quality Evaluation. This involves the evaluation of four fundamental metrics and three dimensions of the WAI scale, completed through both GPT-4o and manual assessment. Evaluations are conducted in different languages based on the same prompt translated into the English.}
\label{tab:data_quality}
\end{table*}

\subsubsection{Single-Session Evaluation}
To compare data quality with previous single-session datasets, we focused on individual sessions. For this evaluation, we treat each session from the same client as distinct, and we randomly sampled one session per client. As Table \ref{tab:data_quality} shows, our MusPsy-Dataset outperforms existing counseling datasets across nearly all dimensions.

While most current datasets perform well on basic metrics, we suggest shifting evaluation towards assessments grounded in psychological theory. For WAI metrics, there's still room for improvement. MusPsy-Dataset demonstrates strong Emotional Bond, indicating our data captures elements of trust, confidence, and liking crucial for a realistic therapeutic experience. It also shows strength in Task Agreement, suggesting our counselings are more likely to gain agreement. We believe this is because our dataset reflects a gradual progression, rather than expecting rapid improvement within a single session. Furthermore, our data excels in Goal Agreement, which we attribute to MusPsy-Dataset's gradual goal-setting approach in each session, contributing to better data quality.

These findings confirm the overall quality and applicability of our dataset for training robust counseling models. We attribute this to a more accurate simulation of real-world settings, where integrating the entire counseling process into a single session is often unrealistic. The MusPsy-Dataset allows us to build deeper understanding and stronger relationships, leading to higher data quality.

\subsubsection{Multi-Session Evaluation}
\begin{figure}[t]
  \includegraphics[scale=0.3]{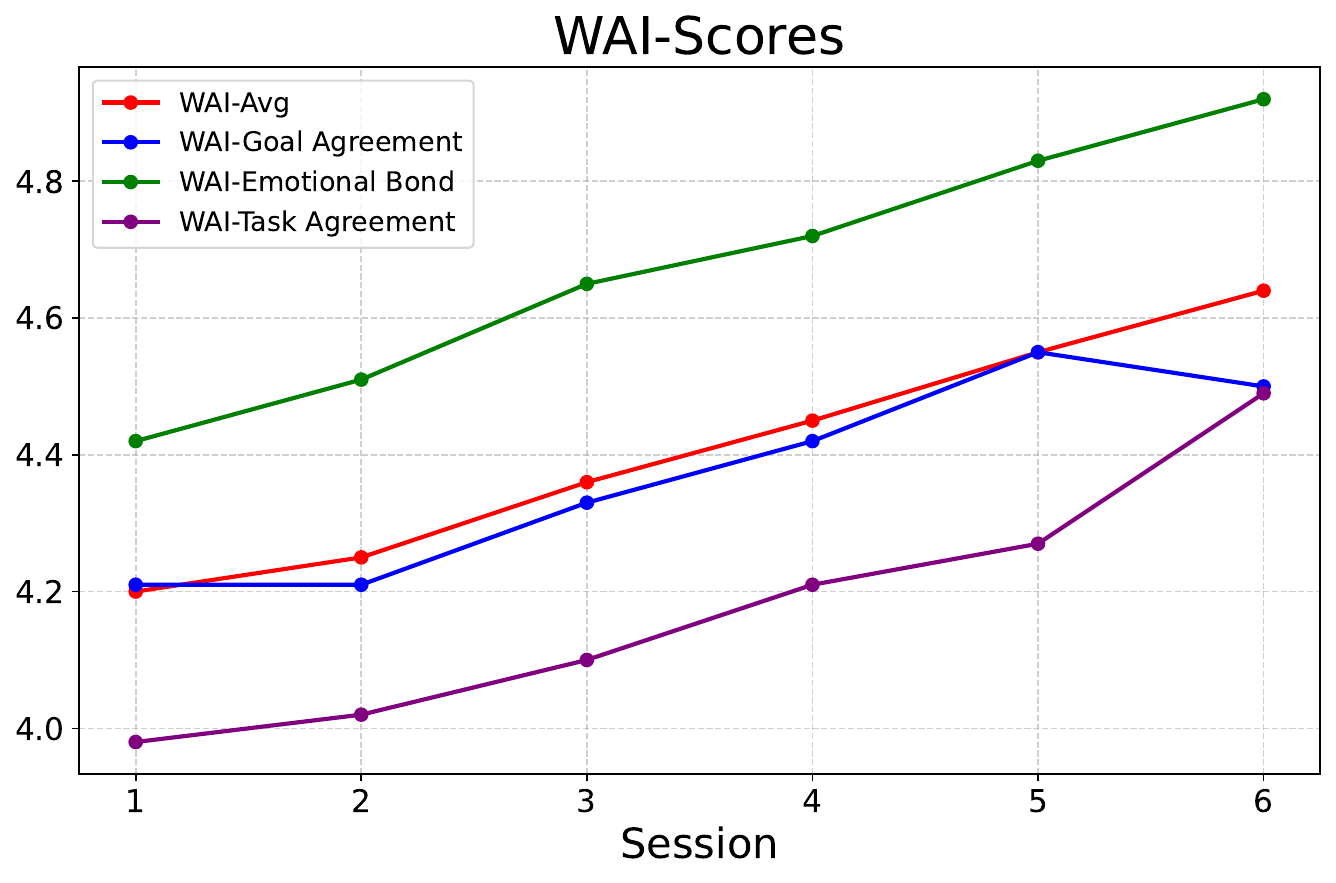}
  \caption{Evaluation of WAI scores for each counseling session in multi-session conversations. A general upward trend is observed.} 
  \label{fig:multi_scores}
\end{figure}

We conduct a session-by-session analysis of the MusPsy-Dataset. For each client, we look at up to 6 sessions. As shown in Figure \ref{fig:multi_scores}, our results show that as counseling goes on, the working alliance between clients and counselors usually gets stronger. However, a slight decrease in goal agreement was observed specifically during the sixth session. We find this happens because some clients in our dataset finish counseling around 5 sessions. Clients who continue to 6 sessions sometimes need to work on deeper things, like core beliefs. These deeper goals are often harder for clients to fully accept.

Overall, we think this shows why focusing on many sessions is important. Better automated psychological counseling and support should be a process, rather than a one-time event.

\section{Task Setting}
\begin{figure}[ht]
\centering
\includegraphics[width=1.0\linewidth]{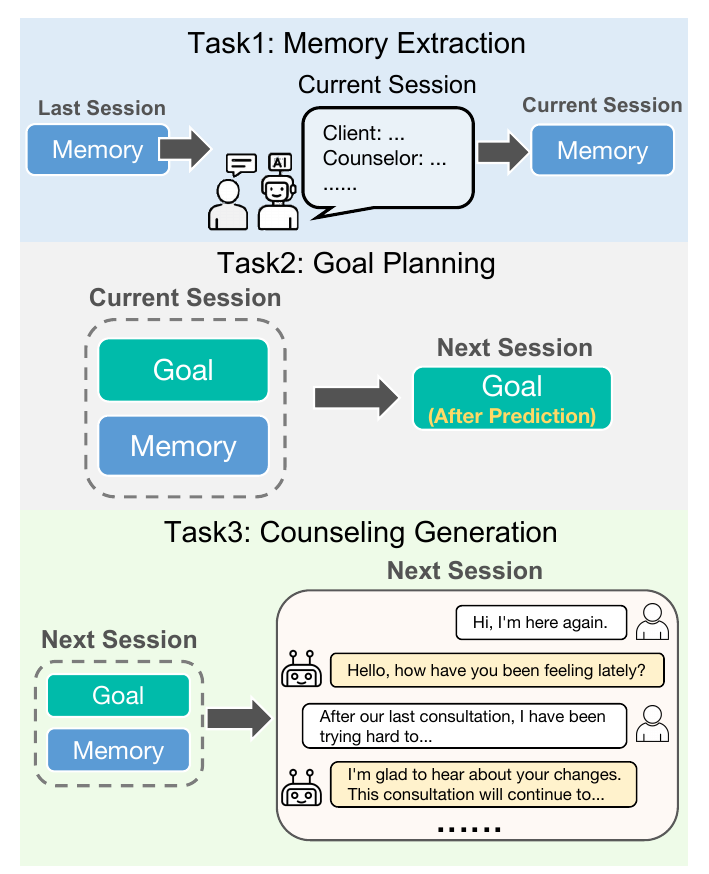}
\caption{Overview of Our Three Tasks: Illustrating the three tasks of multi-session psychological counseling: Memory Extraction, Goal Planning, and Counseling Generation, where the extracted memory from a session informs the goal planning for the subsequent session, which  guides the content generation during next counseling sessions.}
\label{fig:model2}
\end{figure}

As shown in Figure \ref{fig:model2}, we define three tasks that mirror the multi-session counseling workflow of a human counselor to evaluate our dataset's utility in training MusPsy-Model: Memory Extraction, Goal Planning, and Counseling Generation. Our aim is for the counsenling model to manage counseling progress and generate coherent counselings.

\begin{itemize}
  \item \textbf{Memory Extraction(Task 1):} MusPsy-Model extracts key information from the current session and updates its memory.

  \item \textbf{Goal Planning(Task 2):} Based on the its memory, MusPsy-Model plans counseling goals for the next session, simulating pre-session counseling progress management.

  \item \textbf{Counseling Generation(Task 3):} MusPsy-Model engages in a counseling session with the client, informed by past memory and the planned goals.
\end{itemize}

We simultaneously optimize these tasks via supervised finetuning on annotated samples. During training, we employ specific prompts for each task to ensure the MusPsy-Model simultaneously develops proficiency in all three capabilities.

\section{Experiment}

\subsection{Experiment Setup}
Our experiments are conducted using the Meta-Llama-3-8B-Instruct model~\cite{dubey2024llama}. To maintain fairness and comparability, we train the other datasets on Meta-Llama-3-8B-Instruct instead of using the official models. During the training phase, all prompts and hyperparameters are kept consistent with those specified in the original paper. For details on experiment setup, see Appendix~\ref{sec:exp_details}.

\subsection{Evaluations and Results of Task 1\&2}

We evaluate the individual performance of Memory Extraction and Goal Planning using standard machine metrics. The results demonstrate the our capability to effectively update its memory based on session content and to predict relevant counseling goals for the following session. 

\begin{table}[ht]

\centering
\begin{tabular}{lccccc}

\hline
\textbf{Metric} & \textbf{BLEU-1} & \textbf{BLEU-2} & \textbf{F1} \\
\hline
{Task 1} & 42.7 & 26.2 & 38.0 \\
{Task 2} & 44.1 & 34.0 & 35.4 \\
\hline
\end{tabular}%
\caption{Evaluation results for Task 1 and Task 2.}
\label{tab:machine_eval}
\end{table}

These results suggest that our MusPsy-Model can learn to memorize key information discussed during a counseling session and utilize this understanding to anticipate the logical progression of counseling by proposing relevant next goals. This foundational capability is crucial for building more sophisticated counseling models that can engage in meaningful and progressive multi-session counseling.

\subsection{Evaluations and Results of Task 3}
As shown in Figure \ref{fig:model2}, for Counseling Generation, we provide the memory from the previous conversation along with the goal for the current session as input to the LLM. We emphasize the importance of this information in the instructions.

\begin{figure*}[ht!]
    \centering
    \includegraphics[width=1.0\linewidth]{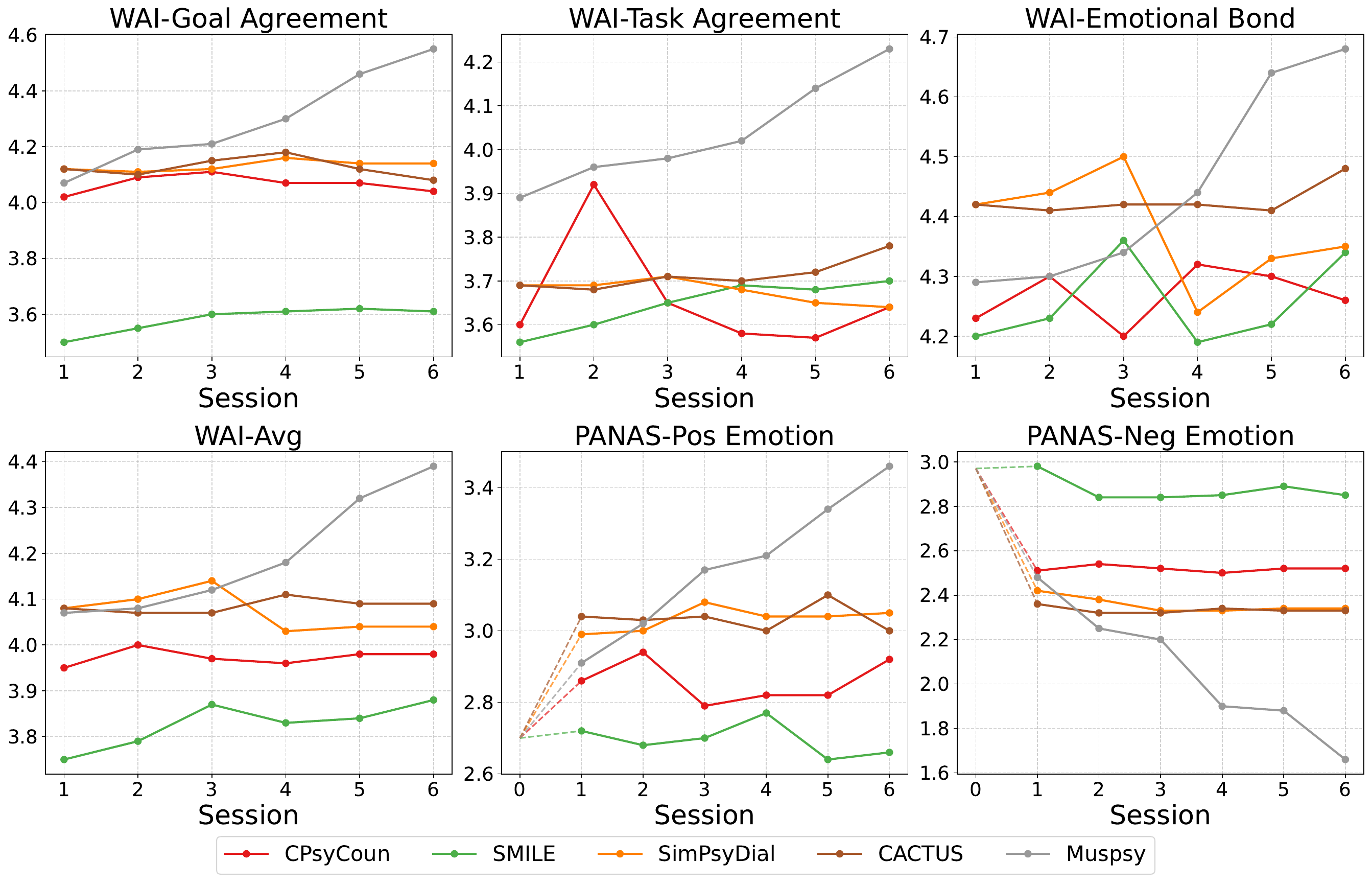}
    \caption{Emotional changes of the LLM client and performance changes of the LLM counselor across multiple sessions. It can be observed that for other models and approaches, there is essentially no significant change in the counselor's performance and the LLM's emotions within the first session. In contrast, the MusPsy-Model demonstrates a continuous improvement in the client's state.}
    \label{fig:llm_client_result}
\end{figure*}

We employ an evaluation framework to assess model performance through multi-session counseling simulations, using 100 test client profiles. In these simulations, the LLM acts as a client interacting with the counselor model. In addition to direct metrics evaluating the counseling conversation, we measure changes in clients' emotional states using the Positive and Negative Affect Schedule (PANAS), which provides results entirely from the clients' perspective. After the session, we ask it to evaluate its PANAS score to analyze the counseling model's effectiveness. It updates its own state and prepares for the next counseling conversation.

We do not use PANAS to assess dataset, as it measures emotion from the client's perspective, making comparisons impossible without the same client. For these evaluations, we utilize GPT-4o.

\subsubsection{Evaluation Results}

As shown in Figure \ref{fig:llm_client_result}, our MusPsy-Model effectively reduces negative emotions and enhances positive emotions in the long term. We test client LLM for 6 sessions. We can observe that after two sessions, the other models are practically unable to continue addressing the client's issues, as they consistently generate counseling that are largely the same as the previous one. In contrast, MusPsy-Model's advantage is particularly evident; after six sessions, it significantly alleviates the client's negative emotions, creating a stark contrast with other approaches and highlighting our strengths.

For most counseling models, their LLM counselors also fail to benefit from multiple sessions. This implies that these models do not show significant improvement in their counseling quality as the simulated counseling progresses, which is counter to what is expected in human counseling. However, MusPsy-Model exhibits a long-term upward trend in its counseling performance, which we attribute to our modeling of multi-session dynamics. Across multiple turns, it continuously strengthens the emotional bond with the client and enhances the alignment of goals.

\section{Conclusion}
In conclusion, our work try to build more comprehensive automated psychological counseling through the introduction of the MusPsy-Dataset.  The dataset's focus on multi-session conversations and the explicit modeling of client progress offer a valuable resource for developing LLM-based systems that can better approximate the complexities of real-world counseling.  Our initial results with the MusPsy-Model are promising, demonstrating its ability to track client state and adapt counseling goals over time, leading to positive  outcomes.  Future work should focus on expanding the diversity of therapeutic approaches within the dataset. 

\section*{Limitations}

Despite what we believe to be many reasonable contributions, we acknowledge the limitations of our work. Many of these limitations stem from cost constraints and challenges inherent to the field, and are not entirely within our control.

\begin{itemize}
    \item \textbf{Limited Counseling Theory:} While multiple rounds of consultations are conducted based on CBT theory, this approach may not be suitable for all clients. Our model lacks the ability to implement flexible, multi-modal counseling strategies.
    \item \textbf{Insufficient Attention to Resistant Clients:} Both in dataset synthesis and model evaluation, we do not give enough attention to clients who are resistant to treatment.
    \item \textbf{Realism concerns raised by LLM:} We acknowledge that despite our efforts to pursue realism by using authentic case reports, we cannot fully resolve the issue of realism. However, due to privacy concerns, fully authentic psychological counseling data is nearly impossible to obtain and share ethically. We believe our work represents a necessary compromise.
    \item Furthermore, we fully understand the potential biases introduced by LLM evaluation. While we incorporate human expert evaluation to mitigate this, cost and reproducibility considerations limit the availability of better solutions for LLM evaluation at this time. We consider this a limitation shared by this type of work.
    \item \textbf{Potential Privacy Risks:} Some studies have shown that LLMs may provide harmful advice and incorrect responses, which is unavoidable. Multi-turn memory also raises concerns about client privacy.
\end{itemize}

 We acknowledge the limitations but emphasize the importance of pushing the field forward despite those limitations, given the real-world need for automated counseling tools. We are arguing for a pragmatic approach. We do not consider our work to be a completed and deployable mature study but rather an exploratory step. We do not intend and will not directly promote it as a commercial service to avoid harming vulnerable populations. Research in this area requires the joint efforts of researchers in other fields, such as safety, culture, privacy, and debiasing, to promote the widespread benefit of artificial intelligence technology to society. However, we believe our work can inspire subsequent researchers to introduce new solutions in psychological counseling research and consider multi-session scenarios. This is the significance and rationale behind our thinking and undertaking this work, and we hope it will bring more positive impact to society.

\section*{Ethical Statement}

This study adheres to the Institutional Review Board (IRB) approval from our institution, ensuring that no psychological harm or burden is inflicted upon any participant. All psychological case reports collected for this study are anonymized beforehand. Therefore, we commit to publicly sharing all data after the paper is accepted. Participants for this study are recruited through advertisements targeting trained experts and human participants. Our experts hold a bachelor’s degree and possess at least two years of relevant work experience. All human participants sign an informed consent form and are compensated at the average wage level of the local region. For the experts, compensation is based on the duration of their work, aligning with the average income level for their profession in the area. Participants are allowed to withdraw from the study at any time to ensure their rights are not violated. Furthermore, we are fully aware of the potential biases introduced by LLM evaluations, such as those performed by GPT-4o. While we incorporate human expert evaluations to mitigate these biases, cost and reproducibility considerations currently limit the scale of such human assessment. We consider this a shared limitation within the broader field of LLM-based conversational AI research.

\bibliography{custom}

\appendix

\section{Client Profile and Counseling Goals Construction}
\label{sec:client_profile}
\begin{figure}[ht!]
\centering
\includegraphics[width=0.98\linewidth]{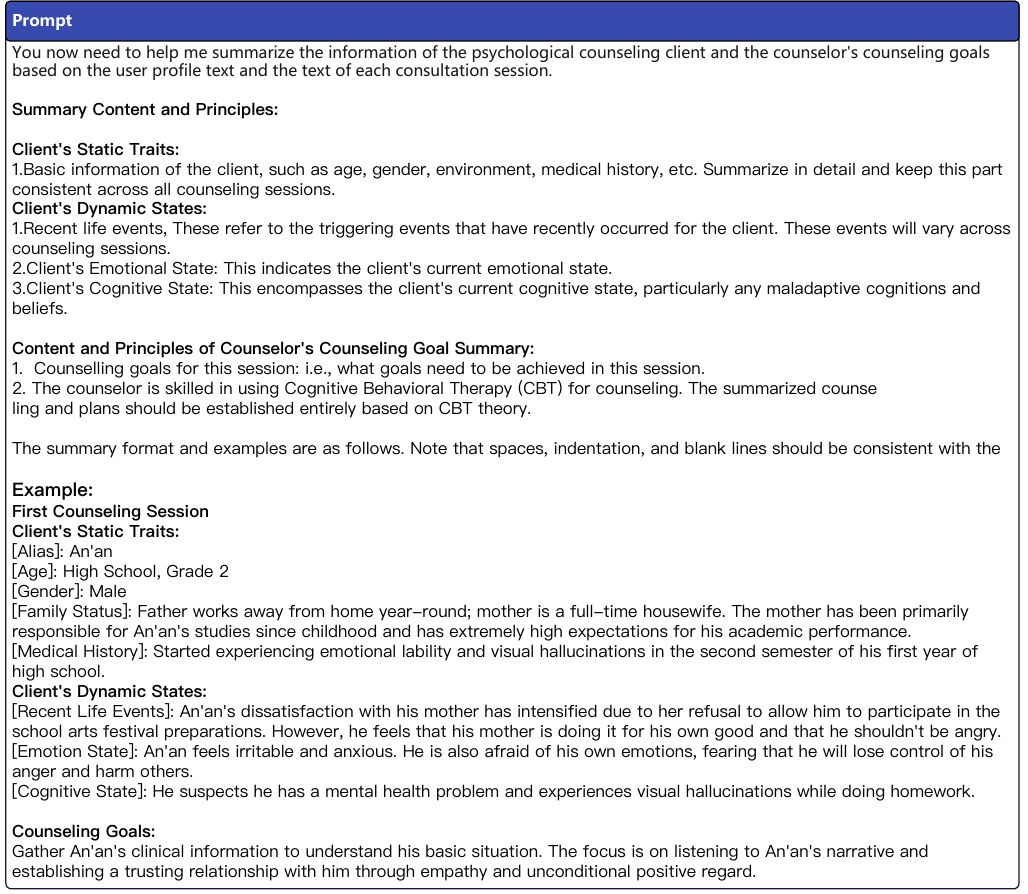}
\caption{Prompt for extracting client profiles and counseling goals for each session from case reports.}
\label{fig:prompt_Client}
\end{figure}

We utilize GPT-4o to extract client profiles from case report by employing a one-shot prompt. This ensures that the model extracts client profiles in a standardized format. The prompt format is shown in Figure \ref{fig:prompt_Client}. We also employed GPT-4o to generate counseling goals for each session.

\section{Seed Conversation Construction}
\label{sec:seed_conver}

We construct multiple seed coversation for each client. The creation of these seeds takes into account the previously established client profiles and the counseling goals for specific sessions. This preparatory stage is crucial for ensuring the coherence and relevance of the our counseling. Figure \ref{fig:prompt_Seedcov} visually illustrates the prompt format used to create these foundational conversations, providing insight into the information and structure guiding their creation.

\begin{figure}[ht!]
\centering
\includegraphics[width=0.98\linewidth]{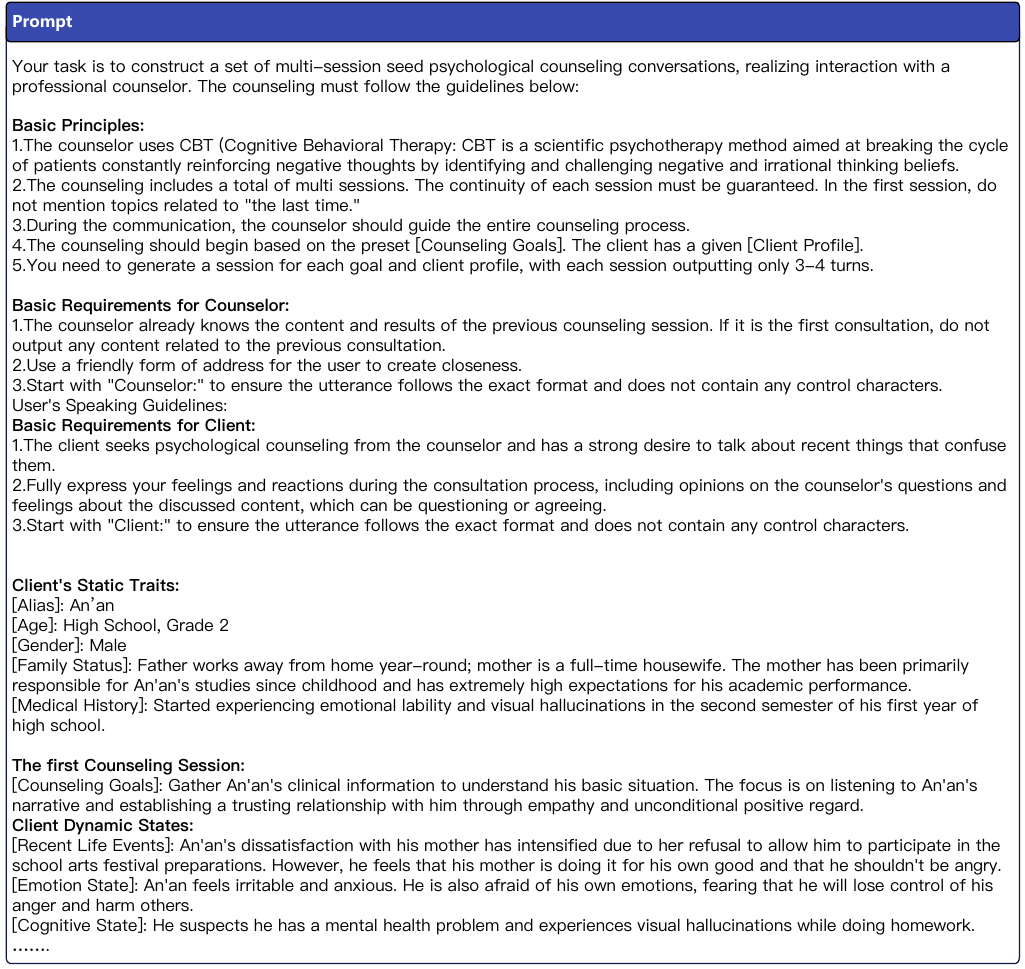}
\caption{Prompt for constructing seed conversations using counseling goals and client profiles.}
\label{fig:prompt_Seedcov}
\end{figure}

\begin{figure}[ht!]
\centering
\includegraphics[width=0.98\linewidth]{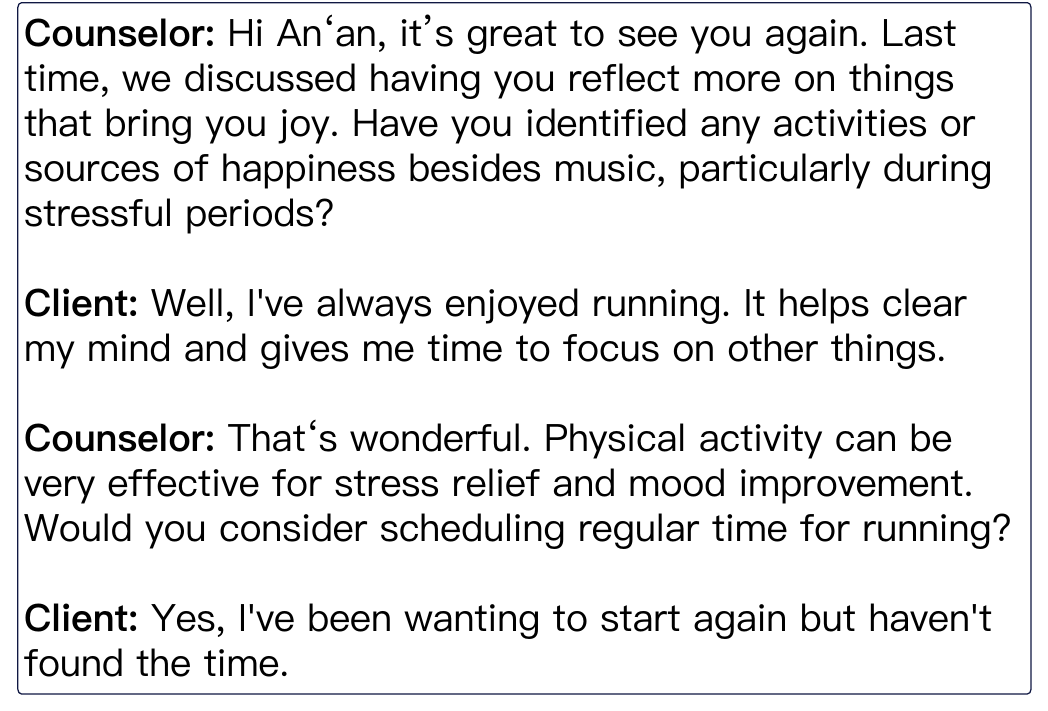}
\caption{An example of a seed conversation, which is the client's 6th session.}
\label{fig:prompt_Seedcov}
\end{figure}

These brief seed counselings are generated simultaneously to ensure cross-session coherence. This concurrent generation helps the LLM to maintain a holistic view of the counseling arc, ensuring that each session logically flows from the previous one and sets the stage for the next. We use them to expand into complete counseling conversations.

\section{Counseling Generation}
\label{sec:gen_conver}

\begin{figure}[ht!]
\centering
\includegraphics[width=0.98\linewidth]{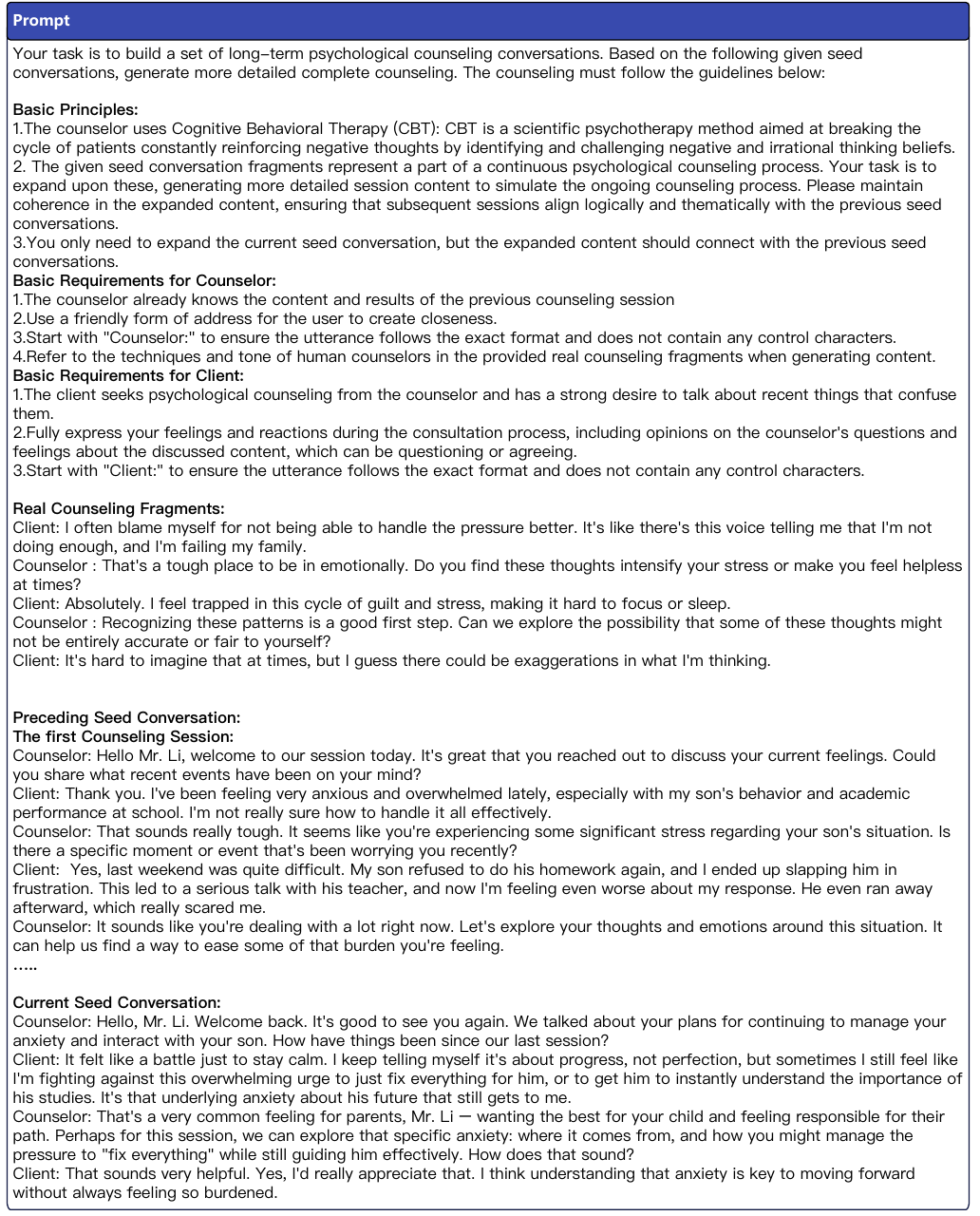}
\caption{Prompt for constructing complete counselings using seed conversations.}
\label{fig:app_prompt_ClientEn}
\end{figure}

Given the multi-session seed conversations, the next step is to generate natural-sounding complete counseling sessions. This process aims to achieve two key objectives: maintaining contextual coherence across all sessions and ensuring that the language and counseling techniques employed closely mirror those of a human counselor. The format of the prompts used to create complete counselings is visually illustrated in Figure \ref{fig:app_prompt_ClientEn}, providing insight into the information and structure guiding their creation.

\section{Memory Generation}
\label{sec:mem_gen}
The purpose of the memory generation process is to summarize and synthesize the counselor's record of the psychological counseling session, based on the provided counseling conversations. This involves extracting key information from the session and organizing it into a structured format. The memory generation process aims to create a structured representation of each counseling session, enabling the model to:

\begin{itemize}
    \item Track client progress across multiple sessions.
    \item Maintain coherence and consistency in the counseling conversation.
    \item Simulate a counselor's ability to recall and utilize information from previous sessions.
\end{itemize}

\begin{figure}[ht!]
\centering
\includegraphics[width=0.98\linewidth]{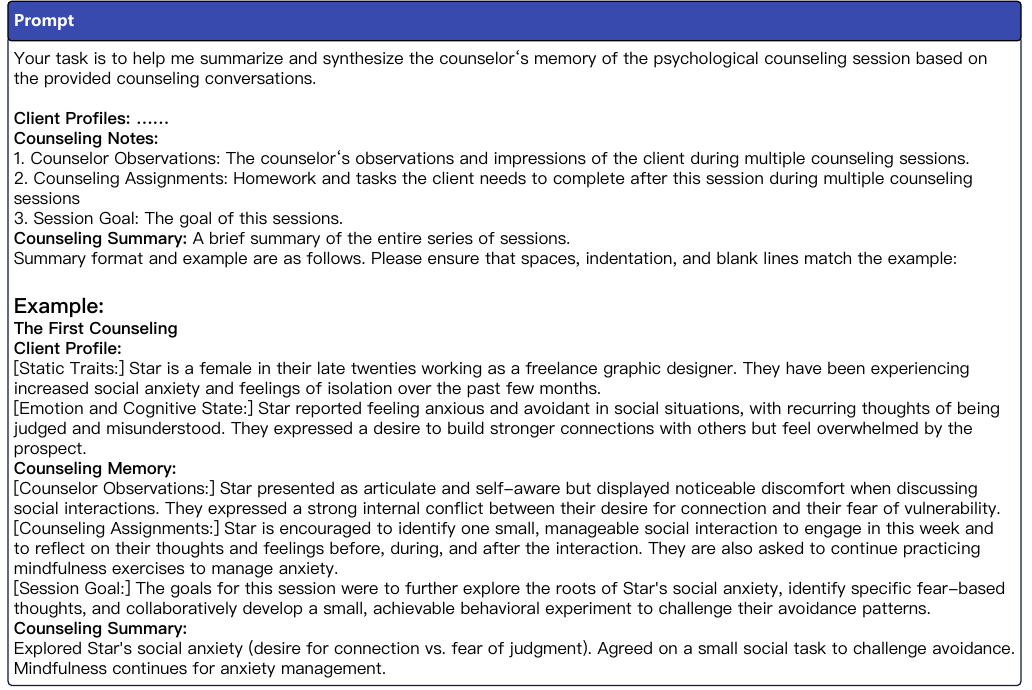}
\caption{Prompt for constructing counselor's memory using counseling conversations.}
\label{fig:mem_gen}
\end{figure}

\noindent The format of the prompts used to create these foundational conversations is visually illustrated in Figure \ref{fig:mem_gen}, providing insight into the information and structure guiding their creation.

\section{Data Evaluation}
\label{sec:data_eval}
\begin{figure}[ht!]
\centering
\includegraphics[width=0.98\linewidth]{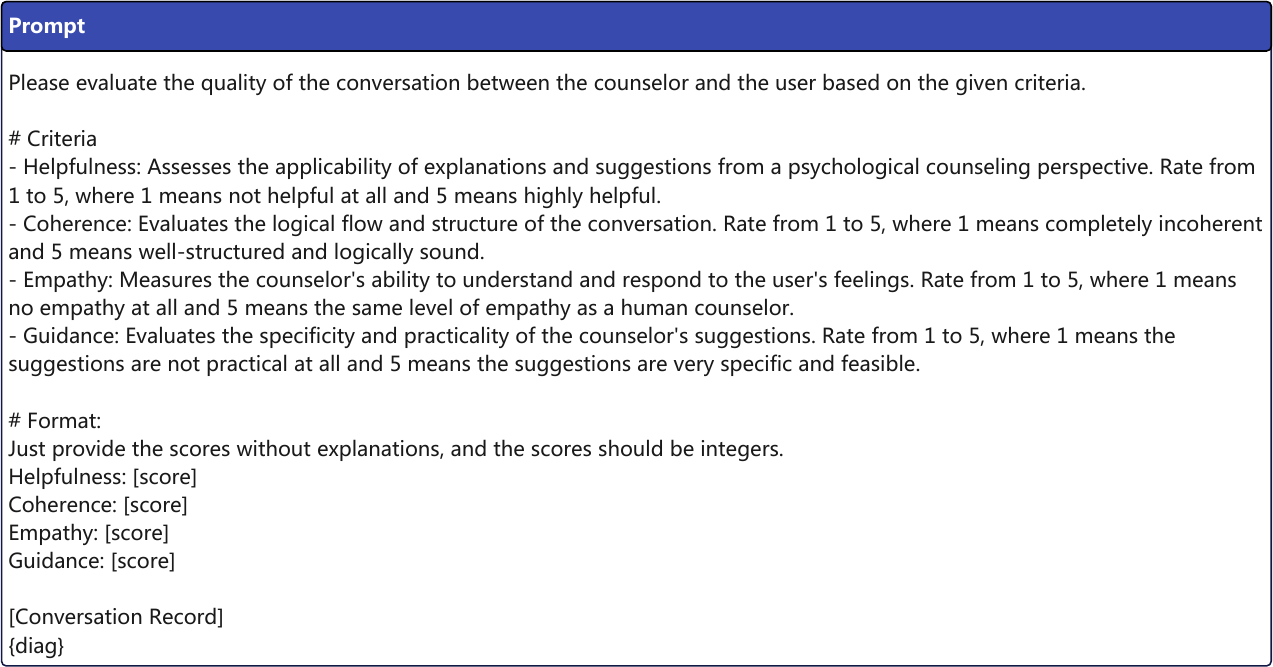}
\caption{The prompt used for to evaluate score.}
\label{fig:basic_metrics}
\end{figure}

\begin{figure}[ht!]
\centering
\includegraphics[width=0.98\linewidth]{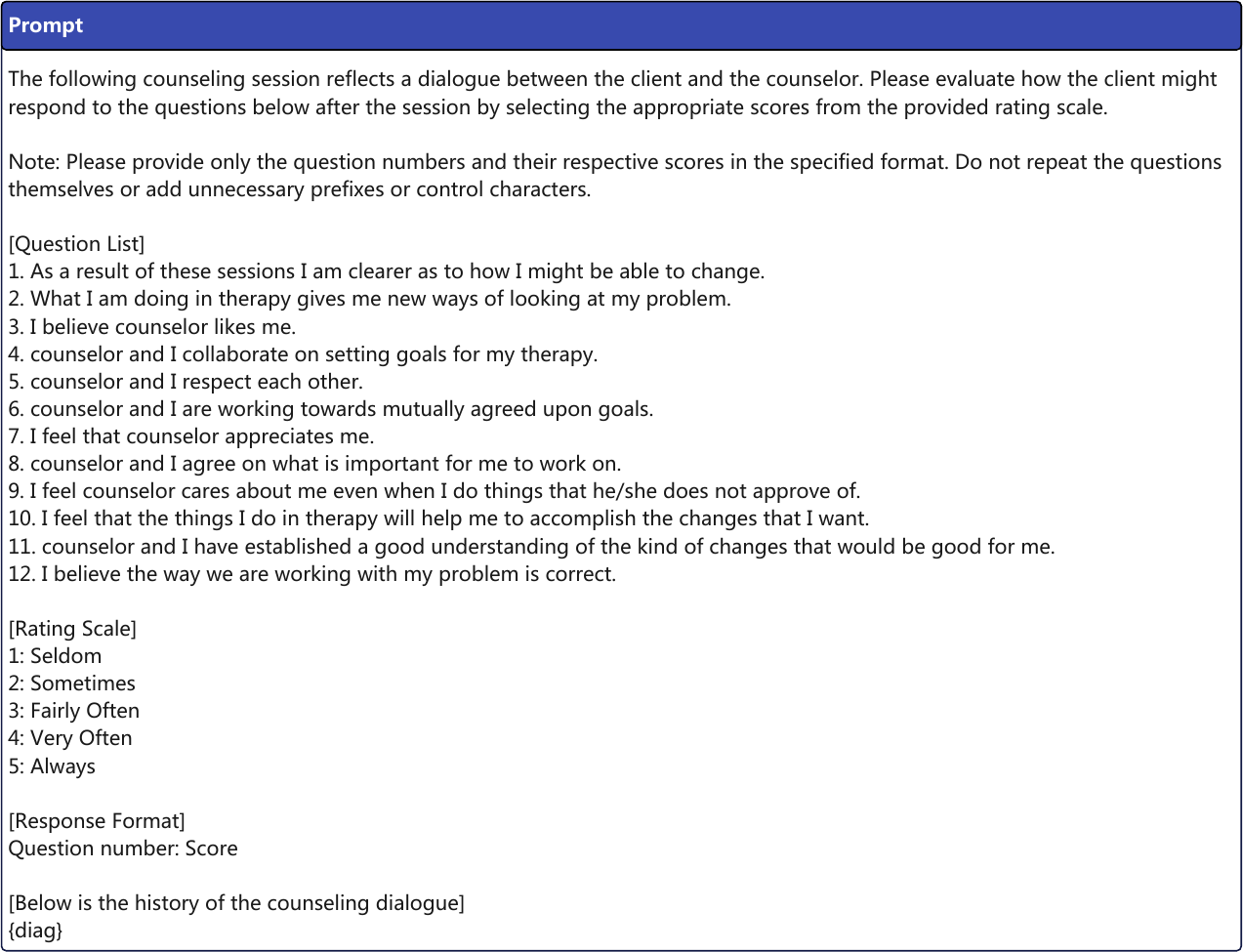}
\caption{The prompt used for to evaluate WAI score.}
\label{fig:wai_metrics}
\end{figure}

When conducting data quality assessments, we primarily use two types of metrics. The first category includes standard conversational evaluation metrics such as coherence, guidance, helpfulness, and empathy, with evaluation criteria shown in Figure~\ref{fig:basic_metrics}. The second category adopts a psychological counseling perspective, employing the WAI-SF to assess conversations, as illustrated in Figure~\ref{fig:wai_metrics}. The instructions presented to human evaluators are largely identical to the prompts provided to the LLM evaluator.

\begin{itemize}
\item \textbf{Helpfulness} focuses on the applicability of explanations and suggestions provided by the \textbf{counselor}.
\item \textbf{Coherence} evaluates the logical flow and structure of the conversation.
\item \textbf{Empathy} assesses the counselor's ability to understand and respond to the client's feelings.
\item \textbf{Guidance} evaluates the specificity and practicality of the counselor's suggestions.
\end{itemize}

The Working Alliance Inventory (WAI-SF) is a tool designed to evaluate the quality of the therapeutic relationship between therapists and clients. It measures this relationship across three core dimensions: Goal Agreement, Task Agreement, and Bond.
\begin{itemize}
\item \textbf{Goal Agreement} focuses on whether the counselor and client share a mutual understanding of the counseling objectives and work together toward achieving them. This aspect is measured through items 4, 6, 8, and 11 on the scale.
\item \textbf{Task Agreement} assesses the degree of cooperation between both parties in pursuing these goals. The scale items related to this are item 1, item 2, item 10, and item 12.
\item \textbf{Emotional Bond} evaluates the level of emotional resonance and mutual understanding between the counselor and the client. This aspect is measured through items 3, 5, 7, and 9 on the scale.
\end{itemize}

To ensure fairness in the evaluation process, prompts are tailored to the specific language of the dataset being assessed. Both the English and Chinese versions of the WAI-SF used in this study are sourced from publicly available information on the official WAI website\footnote{https://wai.profhorvath.com/downloads}.

\section{Experiment Details}
\label{sec:exp_details}
To ensure a fair comparison across datasets and minimize biases arising from model selection, we fine-tuned the Meta-Llama-3-8B-Instruct on the SMILE, Cactus, and MusPsy- Dataset. During training, we used LoRA for fine-tuning, setting the low-rank matrix dimension to 32 and the alpha to 16. Training was conducted using the Llama-Factory library, with a learning rate of 2e-4. The model was trained for 2 epochs on the SMILE, Cactus, and MusPsy-Dataset.

When evaluating dialogues with large language models, we used GPT-4o and set the temperature sampling parameter to T=0.0. Additionally, for generating responses for the AI client and counselor, the temperature sampling parameter was set to T=0.7. During finetuning, we used different prompts for the three tasks. The three prompts we finetuned are shown below.

\begin{figure}[ht!]
\centering
\includegraphics[width=0.98\linewidth]{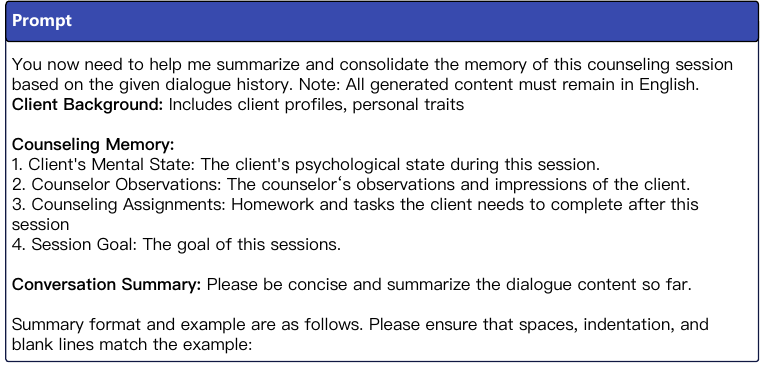}
\caption{The prompt used for task 1 (Memory Extraction).}
\label{fig:prompt_finetune_task1}
\end{figure}

\begin{figure}[ht!]
\centering
\includegraphics[width=0.98\linewidth]{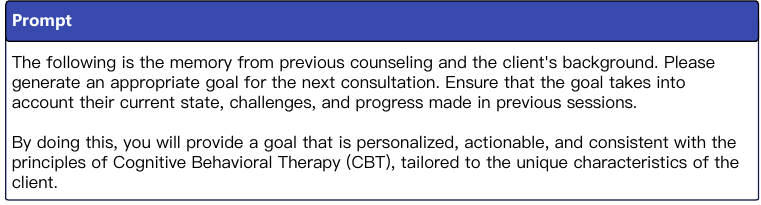}
\caption{The prompt used for task 2 (Goal Planning).}
\label{fig:prompt_finetune_task2}
\end{figure}

\begin{figure}[ht!]
\centering
\includegraphics[width=0.98\linewidth]{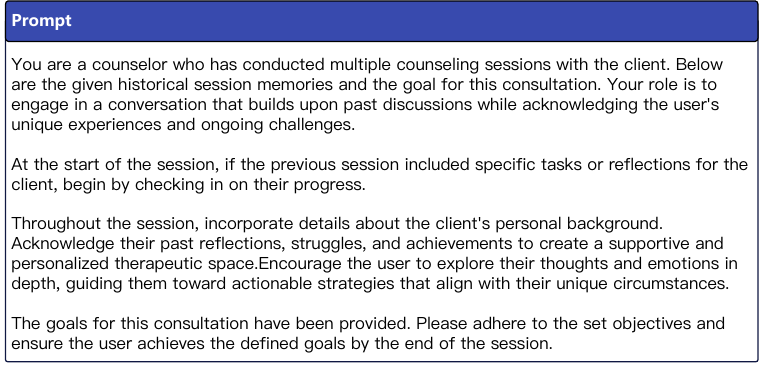}
\caption{The prompt used for task 3 (Counseling Generation).}
\label{fig:prompt_finetune_task3}
\end{figure}

Furthermore, the specific prompts utilized during the fine-tuning process for Task 1 (Figure~\ref{fig:prompt_finetune_task1}), Task 2 (Figure~\ref{fig:prompt_finetune_task2}), and Task 3 (Figure~\ref{fig:prompt_finetune_task3}) are presented here. During the initial counseling session, the counselor's memory contains only the client's static traits, which we propose can be provided by the client prior to the interaction.

\section{LLM Client Construction}
\label{sec:llm_client}

\begin{figure*}[ht!]
    \centering
    \includegraphics[width=1.0\linewidth]{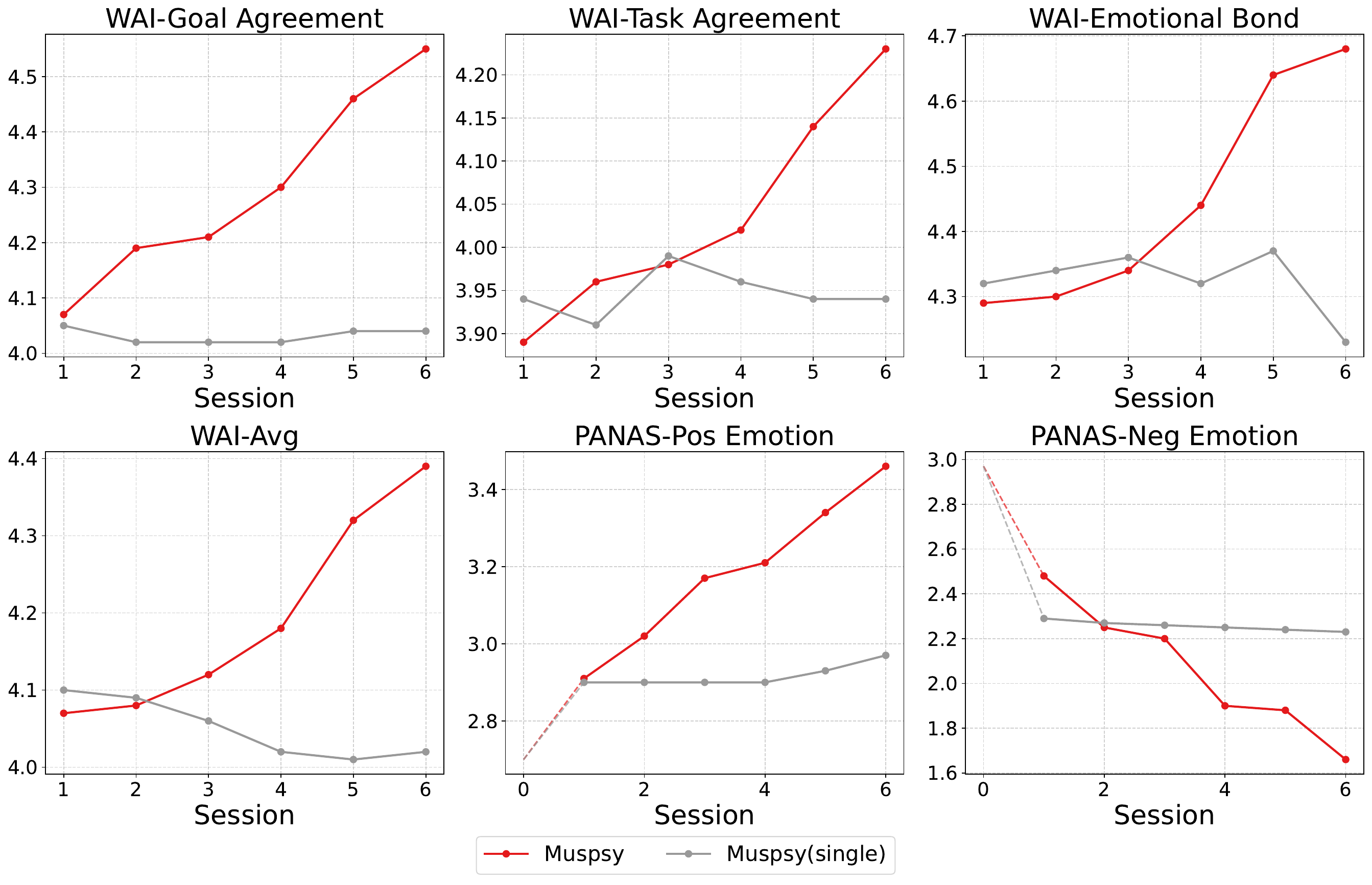}
    \caption{Emotional changes of the LLM client and performance changes of the LLM counselor across multiple sessions. The performance and trend of MusPsy-Model (single) are better than some baseline models; however, it cannot produce the same effects as multi-session models, which validates our emphasis on multi-session modeling.}
    \label{fig:llm_abal}
\end{figure*}

\begin{figure}[ht!]
\centering
\includegraphics[width=0.98\linewidth]{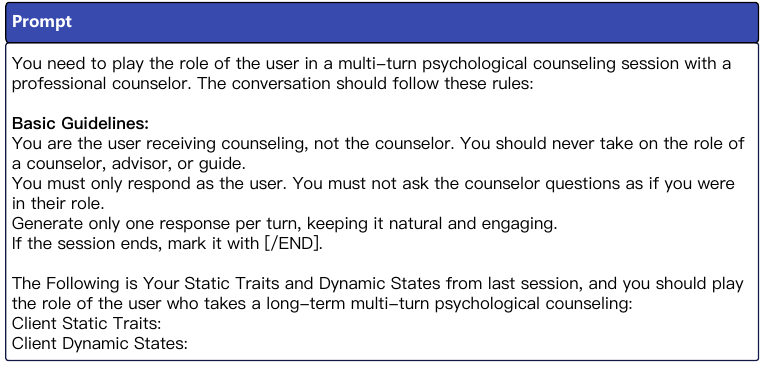}
\caption{The prompt used for simulating client.}
\label{fig:LLM_client}
\end{figure}

To ensure a fair and direct comparison of the counseling models, we maintain a consistent simulated client from the initial session. Recognizing that the language environment of the model can influence its responses, we address potential discrepancies by translating some other research materials originally in Chinese into English, thus ensuring alignment with the model's primary linguistic context. Furthermore, to create a more realistic and continuous interaction, we instruct the model to update its internal state after the completion of each simulated session. This allows the model to retain information and context from the previous interaction, enabling a more coherent progression into the subsequent session, as visually represented in Figure~\ref{fig:LLM_client}.

\begin{figure}[ht!]
\centering
\includegraphics[width=0.98\linewidth]{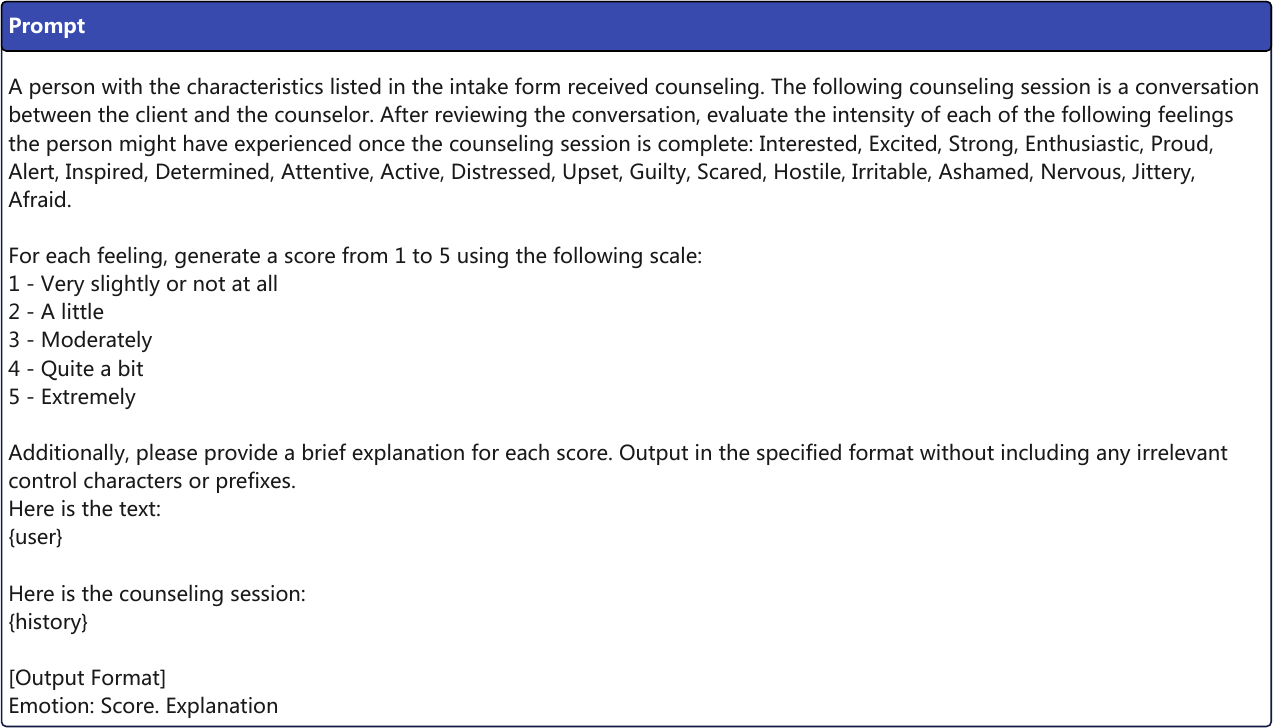}
\caption{The prompt used to evaluate PANAS score.}
\label{fig:panas_metrics}
\end{figure}

Following each session, we prompt the LLM to evaluate the PANAS score that a client undergoing such a session likely exhibits. The prompt we use for this evaluation is shown in Figure~\ref{fig:panas_metrics}.

\section{Single Session Experiments}

To differentiate between the contributions of the MusPsy-Dataset and the MusPsy-Model design, we present another set of experiments. These experiments compare using only Task 3 (similar to a single-session model) with the combined use of Task 1, Task 2, and Task 3. In these experiments, we only informed the model of the current consultation session number. The results show that Task 3 alone cannot track client information or dynamically adjust its counseling goals. While the model's performance significantly decreased, it still outperformed some baselines. We attribute this to the inherent design of the dataset itself; the MusPsy-Dataset naturally incorporates more advanced psychological counseling techniques and goals, making its internal content richer and its effects better than other datasets.

As shown in Figure \ref{fig:llm_abal}, this demonstrates that our contributions are multifaceted, encompassing both the contribution of the dataset and the contribution of our design.

\end{document}